\documentclass[letterpaper, 10 pt, conference]{ieeeconf}  

\IEEEoverridecommandlockouts                              

\usepackage{amssymb}
\usepackage{amsmath}
\usepackage{algorithm}
\usepackage{algorithmic}
\usepackage{tikz}
\usepackage{booktabs}                 
\usetikzlibrary{arrows}
\usetikzlibrary{shapes}
\usepackage{epstopdf}
\usepackage{multirow}

\usepackage{tikz}
\usepackage{textcomp}
\usepackage{hyperref}

\newcommand\copyrighttext{%
  \footnotesize \textcopyright 2019 IEEE. Personal use of this material is permitted.  Permission from IEEE must be obtained for all other uses, in any current or future media, including reprinting/republishing this material for advertising or promotional purposes, creating new collective works, for resale or redistribution to servers or lists, or reuse of any copyrighted component of this work in other works.
  }
\newcommand\copyrightnotice{%
\begin{tikzpicture}[remember picture,overlay]
\node[anchor=south,yshift=10pt] at (current page.south) {\fbox{\parbox{\dimexpr\textwidth-\fboxsep-\fboxrule\relax}{\copyrighttext}}};
\end{tikzpicture}%
}

\renewcommand{\vec}[1]{\mathbf{#1}}

\title{\LARGE \bf
Machine Learning based Simulation Optimisation for Trailer Management  
}

\author{Dylan Rijnen$^{1}$, Jason Rhuggenaath$^{2}$, Paulo Roberto de Oliveira da Costa$^{2}$ and Yingqian Zhang$^{2}$ 
\thanks{$^{1}$ AB InBev, Belgium.
        {\tt\small dylan.rijnen@ab-inbev.com}}%
\thanks{$^{2}$The School of Industrial Engineering, Eindhoven University of Technology, The Netherlands.
        {\tt\small \{j.s.rhuggenaath, p.r.d.oliveira.da.costa, yqzhang\}@tue.nl}}%
}

\begin{document}

\maketitle
\copyrightnotice

\thispagestyle{empty}
\pagestyle{empty}

\begin{abstract}
In many situations, simulation models are developed to handle complex real-world business optimisation problems.
For example, a discrete-event simulation model is used to simulate the trailer management process in a big Fast-Moving Consumer Goods company. To address the problem of finding suitable inputs to this simulator for optimising fleet configuration, we propose a simulation optimisation approach in this paper. The simulation optimisation model combines a metaheuristic search (genetic algorithm), with an approximation model filter (feed-forward neural network) to optimise the parameter configuration of the simulation model. 
We introduce an ensure probability that overrules the rejection of potential solutions by the approximation model and we demonstrate its effectiveness. In addition, we evaluate the impact of the parameters of the optimisation model on its effectiveness and show the parameters such as population size, filter threshold, and mutation probability can have a significant impact on the overall optimisation performance. 
Moreover, we compare the proposed method with a single global approximation model approach and a random-based approach. The results show the effectiveness of our method in terms of computation time and solution quality. 

\end{abstract}

\section{Introduction}


In this paper, we study a trailer management process problem from a big Fast-Moving Consumer
Goods company in Belgium. The company faces problems with managing their transportation trailer 
fleet among the different production sites. Complexities in the trailer management process  include non-homogeneous stochastic demands differing both during the day and during the week, different types of trailers that are used for shipments, the hybrid usage of external and owned trailers and multiple dependent locations managing a shared resource pool with individual needs and preferences. As a result of these complexities, situations occur in which no trailers are available when needed, which negatively impact the logistic processes.  
In order to improve the general trailer management process,  a simulation model was developed that allows analysis of the potential impact of fleet sizing and other trailer management  decisions on the overall process performance. However, finding an optimal configuration of the trailer management process is difficult, as an exhaustive search of the process parameters is infeasible due to the size of the solution space. In this paper, we propose a simulation optimisation (i.e., SO) based approach to optimise the configuration of trailer management process. 

Recent works on simulations with infinite or, finite but having many parameter combinations, have focused on the development of simulation optimisation algorithms that combine metaheuristics with fitness approximation models \cite{Hurrion2000,laguna2002neural,april2003simulation,barton2006,amaran2016simulation}. 
These fitness approximations, also called meta-models are subsequently used to replace simulated fitness evaluations or to select which solutions should undergo a simulation procedure. 

\begin{figure*}[!ht]
    \centering
\resizebox{0.8\textwidth}{!}{%
    \includegraphics{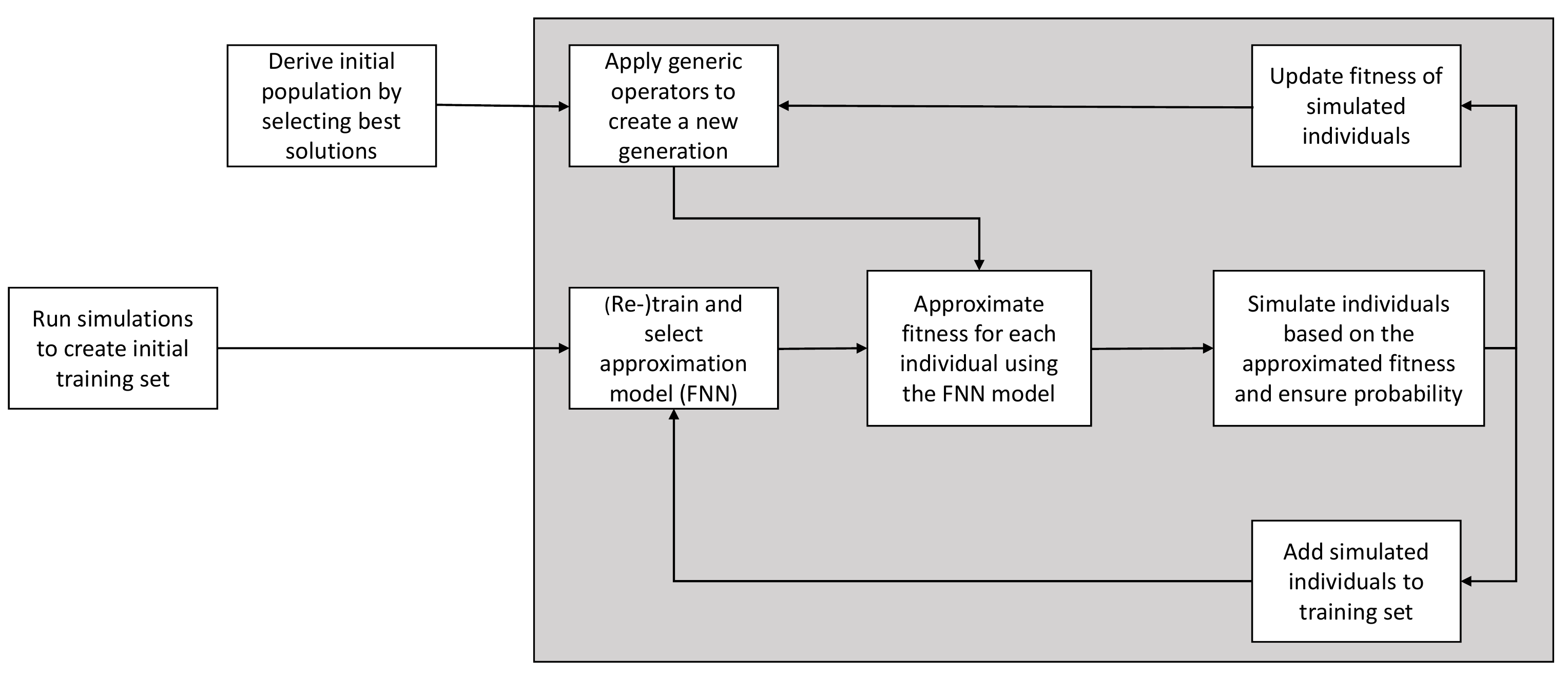}
    }
    \caption{Diagram of the proposed Simulation Optimisation Approach.}
    \label{fig:genhybrid2}
\end{figure*}

In this work, we propose an online simulation optimisation approach (see Figure \ref{fig:genhybrid2}), where a feed-forward neural network is built to approximate the trailer management simulation model, which is used as the meta-model to filter solutions. A genetic algorithm is developed to find improved solutions (i.e., configurations) over iterations. Compared to other approaches of simulation optimisation (e.g. \cite{Hurrion2000,april2003simulation}), our contributions are summarised as below. 
\begin{itemize}
\item While simulation optimisation approaches 
have shown to be effective in various studies, little research has been done on the impact of the parameters of the different components of the optimisation model on the effectiveness of the model. We investigate such impact in this paper. 

\item  
One main concern of the general hybrid simulation optimisation methodology is the lack of convergence property and the relative high chance that optimal solutions get rejected due to statistical noise in the input-output relations of the simulation model and metamodel accuracy \cite{april2003simulation}. In this paper, we consider an \textit{ensure probability} factor that forces evaluation of solutions even when having high approximation fitness scores, which leads to improved performance based on our experimental results.  

\item Few existing works have considered simulation optimisation techniques combined with a machine learning approximation model in supply chain applications using real-world data  \cite{abo2011simulation}. We evaluate the proposed methodology on a real-world trailer management.  
\end{itemize}

The rest of our paper is organised as follows. Section \ref{sec:objfunc} describes the simulation model objective function. In Section \ref{sec:simoptalgo} we present the proposed algorithm and show how the metaheuristic and neural network are combined in the simulation optimisation algorithm. Section \ref{sec:opteval} evaluates the performance of NN in approximating simulation fitness values, and Section \ref{sec:optconfig} evaluates the impact of the parameters of the simulation optimisation algorithm. Lastly, in Section \ref{sec:optapply} we present the results of the simulation optimisation algorithm in our experiments using the trailer management simulation model.

\section{Simulation Model Objective Function}
\label{sec:objfunc}



The objective function aims to minimise the trailer and parking related costs while ensuring trailer and parking availability levels. Trailer costs include the estimated trailer costs of external shipments, the rent cost of owned trailers and the costs of defects. Parking costs include the estimated cost of additional parking places. 

Trailer availability and parking availability can be considered as constraints in the optimisation problem. High unavailability of either trailers or parking results in additional costs in the logistics process and delay other tasks. The simulation model measures the frequency of orders waiting for more than 12 hours to be linked to a trailer. As it can be assumed that linking trailers and orders takes no time, this effectively means that an order could not start processing due to the lack of availability of trailers.

\begin{table}[!ht]
\centering
\caption{Variable notations used in the objective function.}
\label{tab:notation}
\resizebox{1.0\columnwidth}{!}{%
\begin{tabular}{@{}lll@{}}
\toprule
Notation & Description           \\ \midrule
$x^v_{i,l}$                  & Number of vehicles of type $i$ at location $l$                                                      \\
$x^m_{l,i}$                  & \begin{tabular}[c]{@{}l@{}}Multiplication factor for external vehicle arrival rate\\ of vehicles of type $i$ at location $l$\end{tabular}                                               \\
$x^p_{l}$                  & Number of parking places at location $l$                                                   \\

$ x^v_{i}$                  & Number of vehicles of type $i$, i.e. $ \sum_{l} x^v_{i,l}$        \\

$ m^c_{i}$                  & Number of missed calls of vehicle type $i$ \\
$ f^p_{i}$                  & Number of full park occurrences for vehicle type $i$ \\
$ s^e_{i}$                  & Number of shipments on external veh. of type $i$ (dep. on $x^m_{l,i}$) \\
$ d^f$                  & Number of defects \\

$i \in \{1, 2\}$                      & Vehicle type \{1 = Trailers, 2 = Tankers\}  \\

$l \in \{1, 2, 3\}$                      & Location \{1  = Leuven, 2 =  Jupille,  3 = Hoegaarden\}   \\
 \bottomrule
\end{tabular}
}
\end{table}

In the simulation model, constraint violations are modelled via a penalty function (soft constraints) that makes the optimisation problem unconstrained. A linear penalty function was implemented as a cost term in the original objective function \cite{coello2002theoretical}. This penalty function is described in Equation (\ref{eq:costfunction}). In this function $m^c_{i}$ denotes the number of missed calls (or delays) for vehicle type $i$; $c_{mc}$ is the estimated cost for a missed call; $f^p_{i}$ is the number of occurrences of a full parking and $c_{pf}$ is the cost of such occurrence. 

\begin{equation}
\label{eq:costfunction}
    c_{penalty} = \sum_{i} m^c_{i} * c_{mc} + \sum_{i} f^p_{i} * c_{pf} 
\end{equation}
    
Equation (\ref{eq:fitfunc}) shows the complete cost function $f(\vec{x}, s^e_i, d^f, m^c_i, f^p_i)$ for a vector of simulation input parameters $\vec{x}$, composed of variables $x^v_{i,l}$, $x^m_{l,i}$ and $x^p_{l}$, where $i \in \{1, 2\}$ and $l \in  \{1, 2, 3\}$. That is, $f(\vec{x},.)$  is a function of $\vec{x}$ and consists of values for the amount of owned vehicles at each plant, the multiplication factor for external vehicles at each plant and the amount of parking capacity. Since these are the variables considered as input to the proposed model (individuals for the genetic algorithm) we denote $f(\vec{x},.)$ as $f(\vec{x})$ to avoid notation clutter.

\begin{table}[!ht]
\centering
\caption{Cost parameters of the objective function.}
\label{tab:costparams}
\begin{tabular}{@{}lll@{}}
\toprule
Notation     & Description      & Value      \\ \midrule
$c_{own,1}$ & Cost of owned trailer  &2050 \\
$c_{own,2}$ & Cost of owned tanker &1500 \\
$c_{ext,1}$ & Cost of shipment on external trailer&40   \\
$c_{ext,2}$ & Cost of shipment on external tanker&40   \\
$c_{park}$  & Cost of parking &1500 \\
$c_{d}$     & Cost of defect &500  \\
$c_{mc}$    & Cost of delayed order&200  \\
$c_{pf}$    & Cost of parking full &200  \\ \bottomrule
\end{tabular}
\end{table}


Our objective aims to minimise $f(\vec{x})$. In this cost function, $x^v_i$ is the total number of vehicle resources (trailers or tankers) of type $i$. $c_{own,i}$ equals the cost of an owned trailer of type $i$. $s^e_i$ equals the total number of shipments on external vehicles of type $i$ and $c_{ext,i}$ is the cost of a single shipment on an external vehicle of type $i$. The parameter $x^p_l$ equals the amount of parking spots at site $l$ and $c_{park}$ equals the cost per parking spot. The total costs of defects are determined by the number of defects $d^f$ and the estimated cost per defect $c_{d}$. Finally, the penalty cost as derived in Equation \ref{eq:costfunction} is added to the cost function. The description of the notation can be found in Table \ref{tab:notation}.

\begin{equation}
\label{eq:fitfunc}
\begin{split}
f(\vec{x}, s^e_i, d^f, m^c_i, f^p_i) = \sum_{i} (x^v_i * c_{own,i} + s^e_i * c_{ext,i}) \, + \\ \newline \sum_{l} x^p_l * c_{park} + d^f * c_{d} + c_{penalty}
\end{split}
\end{equation}

We note that the cost function incorporates both inputs and outputs of the simulation model. The number of vehicles and parking places are direct inputs in the simulation model while the number of defects, the number of shipments performed on external vehicles and the penalty costs are dependent on the outcomes of the simulation.

The defect and trailer costs were derived from the average historical costs per year. On the other hand, tanker costs were approximated due to the lack of observed historical data. Estimated values were set to slightly lower than trailers costs to reduce their priority in the optimisation process. Delay costs were estimated using expert knowledge, and full parking costs were set to the same level as to have the same importance as delay costs. The complete list of costs used in the objective function can be found in Table \ref{tab:costparams}.

\section{Simulation Optimisation Algorithm}
\label{sec:simoptalgo}
In this section, our proposed approach is presented detailing its three main components: (i) a simulation model; (ii) an approximation model and (iii) a metaheuristic optimisation model. The procedure is outlined in Figure \ref{fig:genhybrid2}. 

The simulation model determines the quality (or fitness) for a given input \(\vec{x}\) via the objective function described in Section \ref{sec:objfunc}. We use the simulation model to create an initial training dataset that contains the fitness for different input values (i.e. different choices for \(\vec{x}\)). This training dataset is used to train a Feed-Forward Neural Network (FNN) model to approximate the simulation model. Finally, we use a Genetic Algorithm (GA) to encode the simulation optimisation problem of choosing the best simulation input values.  Our approach can be considered as an online simulation optimisation algorithm that continuously uses the outputs of the simulation evaluations. The grey box outlines the part of the algorithm that runs continuously until one of the stopping criteria is reached (see Section \ref{sec:ga}). 
We now elaborate on the components (ii) and (iii) of our approach.

\subsection{Genetic Algorithm}
\label{sec:ga}
A GA is used as metaheuristic to search for the best configuration of the input values \( \vec{x}\). Each individual in the population of the GA represents a configuration of the simulation model, where each of the 
genes is one of the decision variables (e.g. a number of vehicles of different types at location $l$ and parking places at $l$).

\paragraph{Initialisation} The initial population is generated as follows. First, a set of simulations is performed on randomly chosen input values \(\vec{x}\).  The inputs and outputs of the simulation are used to train a global approximation model on the solution space. Subsequently, the initial population of the genetic algorithm is created by selecting the input values \(\vec{x}\) with the lowest fitness values.  

\paragraph{Iteration} An iteration of the GA proceeds as follows. Let \( f^{*} \) denote the best fitness value found so far based on the output from the simulation model. During an iteration, the fitness of each individual $\vec{x}$ is approximated using a FNN model. This gives approximations $\hat{f}(\vec{x})$ for every individual $\vec{x}$ in the population. These approximations are checked against the fitness of the best actual simulated solution found so far \(f^{*}\).  The actual fitness for an individual $\vec{x}$, is denoted as $f(\vec{x})$. The approximated solutions $\hat{f}(\vec{x})$ are compared to the best solution so far \( f^{*} \). When the approximation of the solution is smaller than or equal to the best solution so far plus a pre-defined threshold value $d$, i.e., $\hat{f}(\vec{x}) \leq f^{*} + d$ the individual is simulated to get the real output $f(\vec{x})$. Additionally, there is a probability $p_{ensure}$ that an individual will be simulated regardless of its approximated fitness value. This probability is introduced to ensure better convergence of the algorithm, and it is further discussed in Section \ref{sec:impactthreshold}. Whenever a solution is simulated, its inputs and outputs are stored so that they can be used in future retraining of the machine learning model. Subsequently, cross-over (point-wise), elitism and mutation probabilities are applied in order to generate new individuals in the population. 

Based on our experiments, we decided to stop the GA based on a two-rule stopping criterion. The optimisation stops when either 10 generations of the GA have been performed or when 300 simulations have been completed. These criteria were chosen to add a bound to the total available time for optimisation as each generation may result in a number of simulation evaluations equal to the total population size. 


\subsection{Neural Network Approximation Model}
\label{sec:mmtechniques}
Similar to other works in simulation optimisation \cite{Hurrion2000,laguna2002neural,april2003simulation,barton2006} we select a Feed-Forward Neural Network (FNN) due to its capabilities as a universal function approximator \cite{hornik1989multilayer}. The FNN approximation model uses input values $\vec{x}$ of the simulation model (an individual in the genetic algorithm) to predict the value of the cost function $f(\vec{x})$ defined in Section \ref{sec:objfunc}. The subsequent approximations of an individual $\vec{x}$ output approximations $\hat{f}(\vec{x})$. 

During training of the FNN, we normalise the input parameters to allow for better convergence.  We perform grid search using $k$-fold cross validation to select the best performing FNN hyperparameters.  
The number of hidden layers and the level of $L_2$ regularisation on trainable weights are used during grid search. Each neural network is trained using the rectified linear unit (ReLU) as activation functions and has its weights optimised by the Adam \cite{kingma2014adam} algorithm for stochastic gradient updates. After grid search, a final network is trained on the full training set (i.e. $80$\% of the current total dataset). 

\subsection{Simulation Optimisation Procedure}
\label{sec:soprocedure}
After defining the GA and the FNN models we can finally describe the Simulation Optimisation (SO) algorithm procedure: 

\begin{enumerate}
\item Create initial random simulations to gather training data for FNN model as described in Section \ref{sec:mmtechniques}.
\item Create an initial population of size ($S$) consisting of  individuals with the lowest fitness from the random simulations.
\item Train an FNN model using the simulation outputs (Section \ref{sec:mmtechniques}).
\item Apply genetic operators to create a new population.
\item Check for each of these new individuals whether their approximated fitness is lower than the best score found so far plus a threshold value. If so, simulate the individual to get the simulation evaluation, otherwise let the ensure probability decide whether or not the solution will be simulated.
\item At the end of each generation add the simulated individuals with their fitness to the training set and repeat steps 3-6.
\item Continue until one of the stopping criteria has been reached (see Section \ref{sec:ga}).
\end{enumerate}

\section{Neural Network Approximation Model Analysis}
\label{sec:opteval}

The performance of the simulation optimisation method heavily depends on the performance of the approximation model. 
We use two criteria to evaluate the performance of the approximation model: (i) the accuracy of the approximations and (ii) the required amount of training data. 

The accuracy is a necessary measure as inaccurate approximations may lead to inaccurate decisions of the GA.
On the other hand, the required amount of training data is essential as this impacts the effectiveness of the proposed approach. That is, to generate more training data one would need to simulate more input values. In the case of the FNN requiring a large amount of training data to achieve reasonable accuracy, an approximation model would become more costly than running the original simulation model. 


To test the FNN performance, we randomly select \(2000\) configurations for the input values and run the simulation model to obtain the corresponding fitness \(f(\vec{x})\). Next, we create datasets ranging from from size 50 to 2000 in steps of 50 training examples. 
For each dataset, a validation set containing $20$\% of the total examples was separated. The remaining $80$\% was used in the grid search setup (with parameters from Table \ref{tab:fnnparams}) using 10-fold cross-validation to select the best performing hyperparameters.

\begin{table}[!ht]
\centering
\caption{Grid search parameters of the FNN.}
\label{tab:fnnparams}
\begin{tabular}{@{}ll@{}}
\toprule
Parameter                 & Values                                                  \\ \midrule
$\alpha$ ($L_2$ regularisation) & {0.05, 0.01, 0.005, 0.0001, 0.00001}                \\
Hidden layers             & {(15), (25), (50), (50, 5), (50, 25, 15), (25, 15)} \\ \bottomrule
\end{tabular}
\end{table}

The approximation performance is quantified using the coefficient of determination ($R^2$) and the Mean Absolute Error (MAE) metrics in Table \ref{tab:aggmmstats}. We note that, on average, the approximation model can achieve close approximations of the simulation model. Furthermore, when the size of training data is large enough, i.e. more than $500$ samples, most approximation models have an $R^2$ of at least $0.9$ and an MAE lower than $90,000$. With fitness scores ranging between $1,300,000$ and $2,000,000$, this results in MAE errors being less than $10\%$ of the fitness values. We argue that the MAE could form a basis for the threshold of the optimisation algorithm. 
Moreover, in general increasing the dataset size above 500 did not lead to a substantially higher goodness-of-fit.
 

\begin{table}[!ht]
\centering
\caption{Aggregated FNN performance statistics (training set sizes: 500 to 2000)}
\label{tab:aggmmstats}
\begin{tabular}{@{}llll@{}}
\toprule
         & $R^2$ & MAE   \\ \midrule
Mean     & 0.934 & 64,405.1 \\
Std. Dev.& 0.034 & 14,717.0 \\ \bottomrule
\end{tabular}
\end{table}
\section{Simulation Optimisation Parameter Analysis}
\label{sec:optconfig}

In this section we describe the experiments undertaken to define the best performing parameters of the SO method. 

\subsection{Experimental Setup}
\label{sec:expsetup}
The impact of changing different parameters on the optimisation performance is tested using the same set of base parameters for each experiment. The base parameters are designed to use both the threshold and ensure probability mechanics. To ensure comparability, 
the same initial random training simulations are used for all experiments. 
Table \ref{tab:expvalues} shows the base case scenario for the experiments performed.

\begin{table}[!ht]
\centering
\caption{Base case scenario for experimental setup.}
\resizebox{1.0\columnwidth}{!}{%
\label{tab:expvalues}
\begin{tabular}{@{}lllllll@{}}
\toprule
Parameter & \begin{tabular}[c]{@{}l@{}}Population\\ size\end{tabular} & \begin{tabular}[c]{@{}l@{}}Ensure\\ Probability\end{tabular} & Threshold ($d$) & \begin{tabular}[c]{@{}l@{}}Mutation\\ Probability\end{tabular} & \begin{tabular}[c]{@{}l@{}}Cross-over\\ Probability\end{tabular} & 
\begin{tabular}[c]{@{}l@{}}Number of \\ elites\end{tabular} \\ \midrule
Value     & 100                                                       & 0.01                                                         & 100,000   & 0.2 & 0.5                                                           & 2                                                           \\ \bottomrule
\end{tabular}
}
\end{table}

\subsection{Impact of population size}
\label{sec:impactpopsize}

In general, a large population allows for more diversity in the gene pool. However, it also results in a higher number of simulation evaluations. This results in slower running times due to the significant time spent in the simulation model. This effect can potentially cause wasted evaluations on solutions with low fitness. 

To check the impact of the population size, we performed experiments for population sizes ($S$) of $100$, $500$ and $1000$ individuals. We start the SO procedure with $S$ individuals with the best (lowest) fitness and proceed as described in Section \ref{sec:soprocedure}. In Table \ref{tab:parametertuning} the total number of simulation evaluations compared to the best-found fitness are presented. In our experiments, the population size of $S = 100$ showed the best fitness performance at $1,221,940$ and yielded the lowest number of simulation evaluations. We note that it is important to consider that diversity does however not always mean higher population fitness. Instead, our algorithm favours smaller population sizes as it leads to fewer simulation evaluations and exploring more potential solutions.

\subsection{Impact of threshold and ensure probability}
\label{sec:impactthreshold}

In optimisation using genetic algorithms, exploration of the solution space can be achieved using mutation and cross-over operations. However, for our proposed SO method, these operators alone do not necessarily ensure proper exploration. Evaluations are only performed on individuals that have a certain approximation fitness level. That is, what is considered to be a good enough approximated score is dependable on the best fitness found so far. This procedure biases the search for low approximated solutions. Therefore, allowing the model to (re)visit every possible solution is crucial in ensuring a proper search of the solution space.


The threshold parameter is a mechanism to account for the difference between approximated and real simulation output. This parameter influences which solutions undergo simulation evaluation, and it is crucial in achieving convergence. Ideally, the threshold should reduce the required amount of simulations while minimally impacting the convergence of the algorithm. As the threshold is used in collaboration with the best fitness score found thus far, we may have a scenario where solutions with high approximated fitness will never be visited. 
To avoid  only visiting solutions that are similar in fitness quality an \textit{ensure probability} was added to the SO method. During an iteration, mutation operators are used to change the chromosomes of an individual. Thus, there is a probability for each feasible solution to be visited. The ensure probability then forces the evaluation of a low quality an individual to potentially attempt to introduce more population diversity and escape local optima.






The threshold levels were determined based on the accuracy (MAE) of the FNN. 
Experiments were performed for a threshold of $0$ (no threshold), $50,000$ (slightly below MAE), $100,000$ (higher than MAE) and $150,000$ (much higher than MAE). 
In Table \ref{tab:parametertuning} the proposed thresholds, alongside the best-found fitness score and the number of simulation evaluations are presented.
We observe that a lower threshold is beneficial in finding better solutions. The best performance is achieved at threshold $50,000$, but this is only marginally better than the performance of threshold $0$ and $100,000$. A more significant difference exists on the performance of the $150,000$ scenario. Therefore, we conclude that a more restrictive threshold yields lower fitness solutions in less time. Furthermore, using no threshold results in the lowest number of evaluations but also shows a slightly worse performance. This suggests that using the threshold can indeed improve solution quality. 


We compare the impact of the ensure probability by executing experiments using the configuration defined in Table \ref{tab:expvalues} except for $p_{ensure}$. We test probabilities values of: $0$, $0.01$  and $0.1$. 
In Table \ref{tab:parametertuning} the results are shown for each of the probabilities. It can be seen that a small ensure probability will, in fact, improve performance over only using a threshold based approach. We also note that when the ensure probability is higher, the actual performance decreases. This is due to too many unnecessary simulation evaluations which heavily impact the limited simulation budget provided by the stopping criteria, i.e. $300$ evaluations.
 

\subsection{Impact of mutation probability}

Mutation probability impacts the optimisation model by allowing additional exploration of the solution space outside of the existing gene pool. In our experiments three different probabilities: $0.1$ (low), $0.2$ (medium) and $0.3$ (high) were tested. 
In Table \ref{tab:parametertuning} the results are shown for each of the probabilities. It can be seen that a mutation probability of $0.2$ yields the lowest fitness values while performing less than the total amount of simulation evaluations.





\begin{table}[]
\centering
\caption{Best Fitness Score and Number of Simulation Evaluations after parameter tuning.}
\label{tab:parametertuning}
\begin{tabular}{llll}
\toprule
Parameter                             & Values & \begin{tabular}[c]{@{}l@{}}Lowest Fitness \\ Score\end{tabular} & \begin{tabular}[c]{@{}l@{}}Simulation \\ Evaluations\end{tabular}\\ \midrule
\multirow{3}{*}{Population Size}      & 100    & \textbf{1,221,940} & \textbf{256}           \\
                                      & 500    & 1,271,201          & 300                    \\
                                      & 1,000   & 1,250,732          & 300                    \\ \midrule
\multirow{4}{*}{Threshold ($d$)}      & 0      & 1,225,749          & \textbf{140}           \\
                                      & 50,000  & \textbf{1,223,542} & 161                    \\
                                      & 100,000 & 1,224,132          & 263                    \\
                                      & 150,000 & 1,235,804          & 292                    \\ \midrule
\multirow{3}{*}{Ensure Probability}   & 0.0    & 1,232,647          & \textbf{145}           \\
                                      & 0.01   & \textbf{1,225,654} & 257                    \\
                                      & 0.1    & 1,232,839          & 300                    \\ \midrule
\multirow{3}{*}{Mutation Probability} & 0.1    & 1,235,547          & \textbf{125}           \\
                                      & 0.2    & \textbf{1,222,327} & 262                    \\
                                      & 0.3    & 1,252,049          & 300                    \\ \bottomrule
\end{tabular}
\end{table}

\subsection{Comparison against a single global approximation model}

In \cite{barton2006}, the proposed method uses only the machine learning approximations as fitness evaluations in the SO procedure. That is, the algorithm starts by running several simulations and calculating their simulated fitness. This creates a sample of input-output pairs that can subsequently be used for training. Unlike our proposed method, machine learning is not used to filter which solutions to reject. Instead, it completely replaces the original fitness evaluation method. The optimisation is subsequently done only by using the approximated fitness. This effectively eliminates the use of simulation evaluations while running SO. 

We point out that this modification results in no simulation evaluations during the optimisation steps. Thus, it is unknown to the SO whether the population fitness is actually improving. This modification also requires a more accurate approximation model. Therefore, it is often desired that more training data is used during FNN training. 

We compare our SO method to a modified version based on \cite{barton2006}. To achieve the same effect, we prevent prevent any simulation evaluations and FNN retraining during a run. 
We note that this modification saves considerable running time of the SO as simulation evaluations and FNN training are only performed once. In our tests, we train the approximation FNN using $1000$ and $2000$ initial training examples, using the same procedure as shown in Section \ref{sec:soprocedure}.

At the end of the modified SO run, only a limited amount of simulation evaluations are performed. 
Respectively $64$ simulation evaluations in the scenario with $1000$ initial training examples,
 and $24$ simulation evaluations in the scenario with $2000$ initial examples. 
 In our experiments, the best fitness values were respectively $1,253,588$ for 1000 training examples and $1,224,344$ for 2000 training examples. However, the total running time of this modified SO is much longer. 
 The scenario with $2000$ cases would take roughly $11$ hours to generate the training set. Even though the modified SO procedure only takes a couple of minutes in the GA step, the original SO method resulted in similar solution quality while requiring less total running time.




\subsection{Comparison of the optimisation model against random simulation evaluations}
\label{sec:comprand}

As a baseline, we compare the SO to randomly chosen simulation individuals over several runs. In our experiments, we tested random individuals containing $1000$, $1250$, $1500$, $1750$ and $2000$ evaluations. The results showed that the best fitness found over all experiments was $1,337,692$. Moreover, the best fitness score did not improve after $800$ evaluations. When compared to the proposed SO model, regardless of its configuration, it is possible to find multiple better performing individuals using fewer simulation evaluations. Therefore, we argue that the optimisation model offers a significant improvement over a random selection of parameters both in solution quality and running times. 




\section{Results}
\label{sec:optapply}

\subsection{Fitness and running times}

Based on the experiments performed in Section \ref{sec:optconfig}, we report the best performing parameters for the proposed SO algorithm in Table \ref{tab:bestconfig}. We note that the number of elites (i.e. the number of best performing individuals passed from one generation to the next) was not explicitly tested in our experiments. 
We argue that any number of elites should yield a better GA convergence \cite{rudolph1994convergence}. Therefore, we select the number of elites based on the desired running times. Also, we report that the training set size for the initial FNN model contains 1,000 training examples.

\begin{table}[!ht]
\centering
\caption{Best performing parameters of the SO.}
\resizebox{1.0\columnwidth}{!}{%
\label{tab:bestconfig}
\begin{tabular}{@{}llllll@{}}
\toprule
Parameter & \begin{tabular}[c]{@{}l@{}}Population\\ size\end{tabular} & \begin{tabular}[c]{@{}l@{}}Ensure\\ Probability\end{tabular} & Threshold & \begin{tabular}[c]{@{}l@{}}Mutation\\ Probability\end{tabular} & \begin{tabular}[c]{@{}l@{}}Number of \\ elites\end{tabular} \\ \midrule
Value     & 100                                                       & 0.01                                                         & 50,000   & 0.2                                                            & 2                                                           \\ \bottomrule
\end{tabular}
}
\end{table}

\begin{figure}[!ht]
    \centering
    \resizebox{1\columnwidth}{!}{%
    \includegraphics[width=0.8\textwidth]{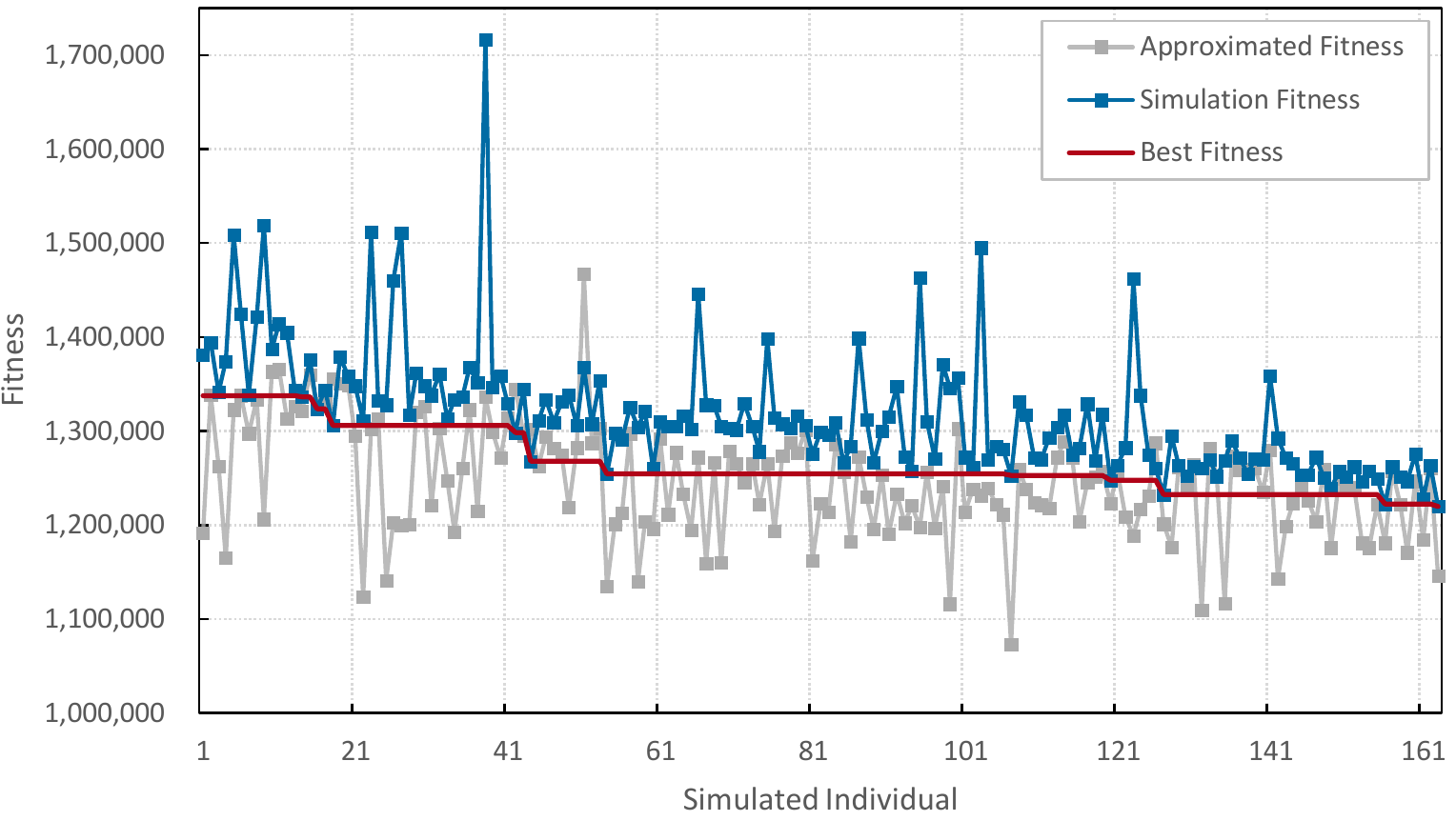}
    }
    \caption{Predicted and real simulation values for the best parameter configuration.}
    \label{fig:bestconfig}
\end{figure}

Figure \ref{fig:bestconfig} depicts the the predicted and real simulation values for the proposed parameters in Table \ref{tab:bestconfig}. The best performing solution $\vec{x}$ selected 110 trailers and 28 tankers in Leuven with multiplication factors $1.061$ and $1.108$ for external trailers and tankers. One hundred thirty-five trailers and two tankers in Jupille, with multiplication factors $1.009$ and $1.180$ for trailers and tankers respectively. Furthermore, $36$ trailers and nine tankers were selected for Hoegaarden with multiplication factor $1.022$ for external trailers. Parking spaces were set to $146$ in Leuven, $92$ in Jupille and $70$ in Hoegaarden. We note that the best solution suggested an increased amount in parking capacity at the different sites when compared to the current situation. This effect is due to the relatively low cost of adding parking places in comparison to the high cost of parking conflicts. The fitness value $f(\vec{x})$ of this configuration equals $1,219,788$. 

The running time required to find this solution can be estimated using the number of simulation evaluations and retraining of the FNN model. Initially, the training set contains $1,000$ simulated examples. After initialisation, $161$ simulation evaluations were performed in the optimisation cycle, and the FNN is trained ten times. In our experiments, one simulation takes approximately $20$ seconds, and one model selection instance takes $2$ minutes. Therefore, the total running time is $407$ minutes (or $6.8$ hours). It is important to note that over $80\%$ of this time is attributed to the generation of the initial training set. Subsequent simulation evaluations consume $15\%$ of the running times while FNN training takes $5\%$ of the total time.

\subsection{Sensitivity Analysis}

We perform a sensitivity analysis on the best-found solution. We increment and decrement each variable in a solution $\vec{x}$ by one and two units. This procedure yields 55 different solutions. We calculate the fitness values $f(\vec{x})$ for each solution and report statistics of the results in Table \ref{tab:sensitivityanalysis}.

\begin{table}[!ht]
\centering
\caption{Summarised statistics of sensitivity analysis for best SO solution.}
\label{tab:sensitivityanalysis}
\begin{tabular}{@{}llll@{}}
\toprule
\begin{tabular}[c]{@{}l@{}}Minimum \\ Fitness\end{tabular} & \begin{tabular}[c]{@{}l@{}}Maximum\\ Fitness\end{tabular} & \begin{tabular}[c]{@{}l@{}}Average\\ Fitness\end{tabular} & \begin{tabular}[c]{@{}l@{}}Standard Deviation\\ Fitness\end{tabular} \\ \midrule
1,216,892                                                    & 1,225,190                                                   & 1,220,241                                                   & 1,547.2                                                             \\ \bottomrule
\end{tabular}
\end{table} 

We note that in Table \ref{tab:sensitivityanalysis}, that the minimum fitness is lower than the solution found by the optimisation algorithm. This result shows that the solution found by the SO is a local optimum. In this specific solution, adding one more tanker vehicle in Hoegaarden decreases the $f(\vec{x})$ value. This change suggests that a local search procedure could improve the SO algorithm. Moreover, we point out that minimum and maximum fitness only differ slightly from the fitness found by the SO method and standard deviation takes very low values. In other words, the solution found by the simulation optimisation algorithm is not susceptible to small changes in the input parameters.

\section{Conclusions}

This paper studies how machine learning can be applied to optimise input parameters for a simulation model on trailer fleet configuration at an FMCG in Belgium. We train a feed-forward neural network to approximate the solutions of the simulator, which is then used to evaluate the fitness values of the genetic algorithm.  We demonstrated the effectiveness of the proposed simulation optimisation method. In the optimised configuration, nearly no trailer unavailability problems persist and parking capacity problems are solved as well. 
The solution found by the SO method reduces the amount of owned tankers and increases the amount of owned trailers. Furthermore, it increases the amount of both external trailers and tankers. The parking is increased at each of the locations to facilitate the extra resources deployed. The higher capacity significantly reduces the trailer availability problems, and the larger parking nearly completely solves the parking problem.

\addtolength{\textheight}{-12cm}   








\bibliographystyle{IEEEtran}
\bibliography{IEEEabrv,thesis}

\end{document}